%% file: 0_main.tex
\documentclass[letterpaper, 10 pt, conference]{ieeeconf}  

\IEEEoverridecommandlockouts                              

\usepackage[backend=biber,
            hyperref=true,
            url=false,
            isbn=false,
            doi=false,
            backref=false,
            style=ieee,
            citestyle=numeric-comp,
            sorting=nyt,
            block=none]{biblatex}

\usepackage{flushend}
\usepackage{balance}

\usepackage{multirow}
\usepackage{amsfonts}
\usepackage{siunitx}
\usepackage{textcomp}
\usepackage{graphicx}
\usepackage{xspace}
\usepackage{hyperref}

\usepackage{amsmath} 
\usepackage{sidecap}

\usepackage{caption}
\usepackage{color,soul}
\usepackage{graphicx}
\usepackage{xspace}
\usepackage{amsmath}
\usepackage{mathtools}
\usepackage{bbm}

\usepackage{algorithm}
\usepackage{algorithmicx, algpseudocode}
\usepackage{wrapfig,lipsum,booktabs}

\newcommand{\algoName}{DAF\xspace}
\newcommand{\algoNameFull}{Deep Affordance Foresight\xspace}
\newcommand{\todo}[1]{\textcolor{red}{TODO: #1}}

\usepackage[dvipsnames]{xcolor}
\renewcommand{\baselinestretch}{0.98}

\title{\LARGE \bf
\algoNameFull:\\ Planning Through What Can Be Done in the Future
}

\author{
    Danfei Xu$^{1}$\thanks{$^{1}$Stanford Vision and Learning Lab, $^{2}$The University of Texas at Austin.}, Ajay Mandlekar$^{1}$, Roberto Mart\'in-Mart\'in$^{1}$, Yuke Zhu$^{2}$, Silvio Savarese$^{1}$, Li Fei-Fei$^{1}$ 
}

\begin{document}

\maketitle
\thispagestyle{empty}
\pagestyle{empty}
\vspace{-20mm}

\begin{abstract}
Planning in realistic environments requires searching in large planning spaces. Affordances are a powerful concept to simplify this search, because they model what actions can be successful in a given situation. However, the classical notion of affordance is not suitable for long horizon planning because it only informs the robot about the immediate outcome of actions instead of what actions are best for achieving a long-term goal. In this paper, we introduce a new affordance representation that enables the robot to reason about the long-term effects of actions through modeling what actions are \emph{afforded in the future}, thereby informing the robot the best actions to take next to achieve a task goal. Based on the new representation, we develop a learning-to-plan method, \algoNameFull (\algoName), that learns partial environment models of affordances of parameterized motor skills through trial-and-error. We evaluate \algoName on two challenging manipulation domains and show that it can effectively learn to carry out multi-step tasks, share learned affordance representations among different tasks, and learn to plan with high-dimensional image inputs. Additional material at \url{https://sites.google.com/stanford.edu/daf}

\end{abstract}

\input{1_intro}
\input{2_related}
\input{3_method}

\input{4_exp}

\input{5_conclusion}

\section{Acknowledgement}
We would like to thank Bokui Shen for helping with
simulation asset creation, and De-An Huang and David Abel
for their insightful feedback and discussions. Danfei Xu acknowledges the support of the Office of Naval Research (ONR MURI grant N00014-16-1-2127). We acknowledge the support
of Toyota Research Institute (“TRI”); this article solely reflects the opinions
and conclusions of its authors and not TRI or any other Toyota entity.

\begin{flushright}
\printbibliography 
\end{flushright}
\end{document}

%% file: 1_intro.tex
\section{Introduction}

Planning for multi-step tasks in real-world domains (e.g., making coffee in a messy kitchen) is a long-standing open problem in robotics. A key challenge is that the tasks require searching for solutions in high-dimensional planning spaces over extended time horizons. An approach to the challenge is to reduce the search problem into a \emph{skill planning} problem: finding a sequence of motor skills applied to objects that will bring the environment to the desired state~\cite{kaelbling2011hierarchical,kaelbling2013integrated,kaelbling2017pre,garrett2020pddlstream,toussaintlogic,toussaint2018differentiable}. However, this reduction leads to a \emph{combinatorial space} of possible skill parameters and object states, which can still be prohibitively expensive to plan with. On the other hand, only a small subset of skills can be carried out successfully at a given state in a typical manipulation domain. 
Thus to be effective, it is crucial for a planner to focus only on skills that are executable in a given environment state.

\begin{figure*}[t]
  \centering
  \includegraphics[width=0.9\linewidth]{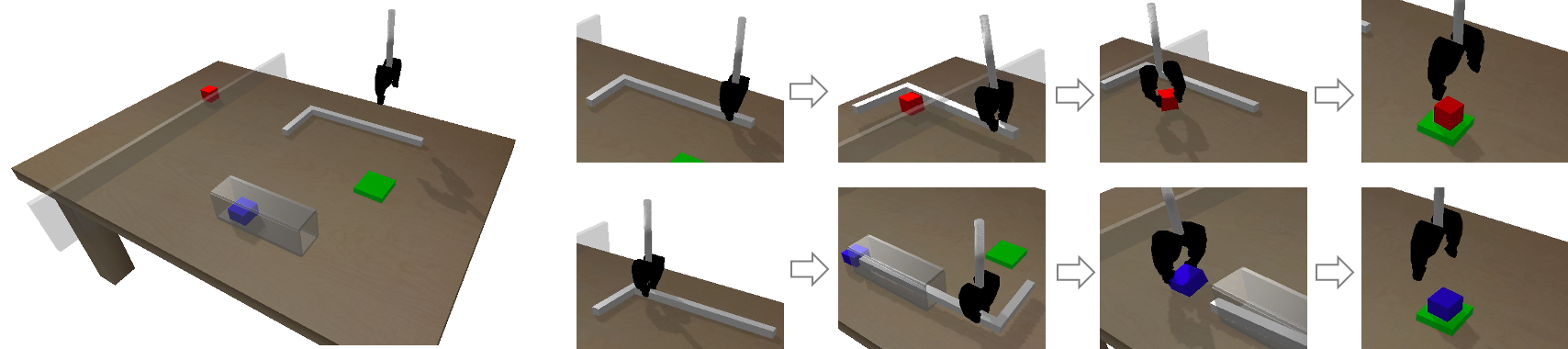}
  \caption{We illustrate our method in a tool-use domain (left). Classical affordance models the \emph{immediate effects} of actions. In this case, a \texttt{graspable(tool)} would only inform that the tool can be grasped and held. However, the two tasks shown on the right require the robot to grasp the tool differently depending on the task goal (red vs. blue on target). In this work, we extend the concept of affordance to represent longer-term effect of actions and enable a robot to learn to reason about what actions it should take next in order to achieve a long-horizon task goal.
  }
  \label{fig:pull}
        \vspace{-5pt}
\end{figure*}

The ability to reason about what actions are possible in a given situation is commonly studied through \emph{affordances}.
Classically, an affordance is the \emph{potential for actions} that an object ``affords'' to an agent~\cite{gibson1977theory}. For example, a mug is ``graspable'' and a door is ``openable''.
These affordances can be refined to consider the exact parameterization of the action that may lead to success. For example, prior works in robotics have used affordance to represent possible grasping poses based on images of objects~\cite{mahler2017dexnet,bousmalis2018using,detry2011learning,mandikal2020dexterous}. However, we argue that this classical notion of affordance is \emph{myopic} and unsuitable for skill planning. This is because an affordance only implies the potential of carrying out an action, ignoring the action's effect on the subsequent plan towards a long-term goal.  
Consider the scene in Fig.~\ref{fig:pull}: a myopic ``graspable'' affordance of the tool only implies that an agent can grasp and hold the tool, but different tasks may require different grasping poses. For example, using the tool to hook the red cube requires a different pose than for pushing the blue cube out of the tube. 

In this work, we propose to use a learned environment dynamics model to extend the concept of affordances to represent the \emph{future actions} that would become feasible if a certain action is executed at the current state, thereby informing the agent the best actions to take to achieve a long-term goal. 
For example, given a task goal of grasping the red cube, we aim to model whether a grasping pose would enable the robot to use the tool to hook the cube. This would subsequently depend on whether an enabled \texttt{hook} action would make a \texttt{grasp(red-cube)} action feasible. 



To develop the method, we adopt a relaxed notion of affordances. Classically, an afforded action is both \textit{feasible} (e.g., robot kinematics allows reaching the target grasping pose) and can \textit{achieve a desired effect} (e.g., the tool being grasped stably). As discussed above, different task goals may require different action effects (e.g., different in-hand poses). 
Instead, we relax the definition of affordance to only model the feasibility of an action, and represent the effect of an action as the \emph{expected affordances at future states}. In other words, we wish to model (1) what actions are feasible at a given state and (2) what actions would become feasible if an action is executed.
This recursive structure allows composing chains of affordances to reason about long-horizon plans. 

Concretely, we introduce \algoNameFull (\algoName), a \emph{learning-to-plan} method that incrementally builds environment models around the affordances of parameterized motor skills~\cite{da2014active}, and learns to plan for multi-step tasks through trial-and-error. 
\algoName learns a latent dynamics model to predict future latent states conditioned on sampled skill plans and an affordance prediction model to evaluate skill affordances both at the current and future latent states. \algoName can use both models together to select multi-step plans that are most likely to achieve a task goal. 
Moreover, \algoName can be trained end-to-end from pixel observations, allowing \algoName to model complex dynamics such as pouring liquid, for which manually defining an affordance is hard.

We present evaluation results on two robotic manipulation domains. In the \emph{Tool-Use} domain as shown in Fig.~\ref{fig:pull}, a free gripper robot must use the hook-like tool to fetch the red and blue cubes and put them on the green target, evaluating the capabilities of our robot to differentiate between the same affordance (graspable) based on future task needs. The \emph{Kitchen} domain requires the robot to plan through complex dynamics such as pouring liquid to complete multi-stage tasks of serving tea or coffee, highlighting the ability of \algoName of combining and reusing learned affordances for other tasks.
We demonstrate that \algoName significantly outperforms a competitive latent-space planning method~\cite{hafner2019learning} and can effectively learn to plan with high-dimensional image input. 

%% file: 2_related.tex
\section{Related Work}
\subsection{Affordance}
Affordances have a rich history in fields such as robotics, psychology, computer vision, and reinforcement learning (RL). In robotics, many have used affordance as a representation prior, e.g., predicting grasping poses~\cite{mahler2017dexnet,bousmalis2018using,detry2011learning,mandikal2020dexterous}, traversable regions~\cite{ugur2007learning}, and exploration~\cite{nagarajan2020learning}. As discussed in the introduction, such a notion of affordance cannot be easily adapted for planning due to its myopic nature. 

Other works have explored learning affordance with respect to a task goal. For example, task-aware grasping~\cite{dang2012semantic,fang2020learning,song2010learning,zeng2018learning} predicts grasping poses in anticipation of a task goal (e.g., tool-use~\cite{fang2020learning}). However, each learned affordance representation is tied to a specific task goal (e.g., a specific way of using the tool). In contrast, our affordance representation can be flexibly composed to reason about diverse long-horizon plans with different task goals.

Theoretical works in RL have formalized affordance for sequential decision making~\cite{abel2014toward,abel2015goal,cruz2016training,khetarpalcan}. Closest to us is Khetarpal \emph{et al.}~\cite{khetarpalcan} that introduces the notion of ``intent''. Intent specifies the desired future state distribution of an afforded action. By modeling the satisfiability of intents, they build partial models of environments that allow efficient planning. A key limitation of this work is that the intents are complex functions that are hand-defined, e.g., a ``Move Left'' intent checks the agent's x-position change in a grid world. Such detailed conditions are tedious and difficult to specify robotics domains. For example, in our \emph{Tool-Use} domain, one would need to define different intents for grasping the tools at different locations. This also requires specifying the precise relative poses between the gripper and the object. In contrast, our affordance represents action feasibilities, which can be checked via robot kinematics or a crude collision detector and shared across all tasks in the same domain. 

\subsection{Task and Motion Planning}
Our definition of affordance is closely related to ``preconditions'' or ``preimages'' in Task and Motion Planning (TAMP)~\cite{kaelbling2011hierarchical,kaelbling2013integrated,garrett2020pddlstream,toussaintlogic,toussaint2018differentiable,garrett2020integrated}. Most TAMP methods require fully-specified planning space and dynamics models. Recent works proposed to build dynamics models by characterizing the preconditions and effects of skills~\cite{kaelbling2017learning,wang2018active,pasula2007learning,xia2018learning}. For example, Kaelbling \emph{et al.}~\cite{kaelbling2017learning} proposes to learn the preimage of a skill given desired effects through trial-and-error.  However, they still require predefined planning spaces such as object poses. Our method plans in a learned latent space with image input. This enables our method to model complex dynamics such as pouring liquid, for which manually designing a planning space would be challenging.

Our planning formulation is heavily inspired by works from Konidaris and colleagues~\cite{konidaris2014constructing,konidaris2015symbol,konidaris2018skills,ames2018learning}, which aim to build compact symbolic environment models by capturing skill pre-condition and effect distributions through interaction. The resulting symbolic models are provably both necessary and sufficient to verify whether a skill plan is \emph{sufficing}~\cite{konidaris2018skills}, meaning that the plan is executable (analogous to plan completion probability defined in Eq.~\eqref{eq:cost}) and leads to a goal (analogous to our \emph{goal-directed plans}). However, these symbolic representations, once built, are confined to a fixed domain. A recent work~\cite{james2019learning} attends to this limitation by building symbols on an agent-egocentric space that facilitates cross-domain generalization. Our work offers a different perspective and propose a latent planning formulation that exploits the generalization ability of deep neural networks. 
Our idea of composing affordance for planning is also related to option chaining~\cite{konidaris2009skill,konidaris2012robot}, although we do not explicitly construct skill trees.

Our method is related to works that learn to predict TAMP plan feasibilities from observations~\cite{wells2019learning,driess2020deepheuristics,driess2020deepreasoning}. For example, Deep Visual Reasoning~\cite{driess2020deepreasoning} learns to generate feasible plan skeletons for Logical Geometric Programming (LGP) solvers. A drawback of these approaches is that they rely on TAMP/LGP planners that can already solve the task to generate planning supervisions for training. In contrast, our method learns through trial-and-error.

\subsection{Learning to Plan}
Our method is related to learning dynamics models for model-based RL~\cite{oh2015action,agrawal2016learning,finn2017deep,hafner2019learning}. Most recent works have focused on building complete environment models directly from the raw observation space. However, learning to make accurate predictions with high-dimensional observations is still challenging~\cite{oh2015action,watter2015embed,finn2017deep}, especially for visually complex long-horizon tasks. Instead, our method builds a \emph{partial model}~\cite{talvitie2009simple} of the environment on skill affordances, which are low-dimensional and amenable to long-horizon planning. 

Prior works on planning with partial models focuse on predicting either reward or quantities that are tied to a task~\cite{dosovitskiy2016learning,oh2017value,amos2018learning,hafner2019learning}. For example, PlaNet~\cite{hafner2019learning} learns to predict future rewards and observations through a latent dynamics model. It is difficult for these methods to share reward models among different tasks and transfer to new tasks. In contrast, our composable affordances are defined independent of a final task goal. This enables our learned affordance and dynamics models to be shared and reused among different tasks to improve data efficiency and task performance. 

%% file: 3_method.tex
\section{Method}
The primary technical contributions of this work are (1) a new form of affordances suitable for multi-step planning and (2) a method for learning to plan with the affordances of parameterized motor skills~\cite{ames2018learning}. Here we first lay out the decision making problem setup, then describe the affordance-based planning problem formulation, and finally present the learning-to-plan method \algoNameFull. 

\subsection{Problem Setup}
We consider partially observable domains with observation space $O$, state space $S$, parameterized skills $\Pi$ (described later), and transition dynamics $\mathcal{T}: S\times \Pi\rightarrow Dist(S)$. We assume a finite set of goals $G$. Each $g \in G$ is a binary condition function $g: S\rightarrow \{0, 1\}$ indicating if a state is in a goal state set $S_g$. The objective is to reach the goal by the end of an episode. 

\input{algo1}

Following prior work~\cite{ames2018learning}, we define a parameterized skill~\cite{da2014active} by a policy $\pi(s, \theta)$ modulated by a set of parameters $\theta \in R^D$. For example, a grasping skill ($\pi$) can be parameterized by 3D grasping positions ($\theta$), and the policy can execute a planned grasping motion. An important feature of motion planning-based skills that we leverage in this work is that we can check if a skill is \emph{feasible to execute} before executing it. The feasibility can be determined through robot kinematic constraints or if a skill motion plan would result in unintended collision between the robot and the environment. For example, in the setup shown in Fig.\ref{fig:pull}, grasping the red cube directly is infeasible due to the kinematic constraint defined by the virtual wall, and grasping the blue cube would collide the gripper with the pipe, which is also infeasible.

Skill feasibility checkers are commonly used to prune skill samples in solving a larger task-and-motion-planning (TAMP) problem~\cite{kaelbling2011hierarchical,kaelbling2013integrated,kaelbling2017pre,garrett2020pddlstream,toussaintlogic,toussaint2018differentiable}. TAMP methods typically require knowledge of ground truth states and an environment dynamics model. Instead, we leverage skill feasibilities to develop a method that can learn to plan in an environment with unknown dynamics. 


\subsection{Planning with Affordances}
Here we formally define our affordance representation and introduce a planning problem setup based on affordances. 

\textbf{Definition 1} (Affordance $\mathcal{A}$): \textit{Given a skill $(\pi, \theta)$, we define an affordance as $\mathcal{A}_{\pi, \theta} = \{s\in S |(\pi, \theta)\text{ is feasible at }s\}$. We use $\mathcal{A}_{\pi, \theta}(s)=\mathbbm{1}[(s, \pi, \theta)\in \mathcal{A}_{\pi, \theta}]$ to denote if state $s$ affords $(\pi, \theta)$. }

To formalize a planning problem using $\mathcal{A}$, we first show how to compute the probability of \emph{plan completion} from some initial state distribution. A length-$N$ plan $p$ belongs to the set $\mathcal{P}_N=\{(\pi_i, \theta_i)\}_{i=1}^N | (\pi_i, \theta_i) \in \Pi, N\in \mathbb{Z}^{+}\}$. A particular plan $p \in \mathcal{P}_N$ is then a sequence of parametrized skills $\{(\pi_1, \theta_1), \ldots, (\pi_N, \theta_N)\}$. Without loss of generality, we assume fixed plan length and omit the subscript $N$.
Given a plan $p$, we denote the induced state distribution at each step $i$ as $Z_i(\cdot; p)$. Given an initial state distribution $Z_0(\cdot; p)$, $Z_{i>0}(\cdot; p)$ can be expressed recursively as:
\begin{align}
    \label{eq:dist}
    Z_i(s'; p) \propto \sum_{s\in S} \mathcal{T}(s'|s, \pi_i, \theta_i) Z_{i-1}(s; p) \mathcal{A}_{\pi_i, \theta_i} (s)
\end{align}
where $(\pi_i, \theta_i)$ is the skill at step $i$ of plan $p$.
We can compute the probability of completing the plan $p$ (being able to execute each skill in the plan) starting from $Z_0$ as:
\begin{align}
    \label{eq:cost}
    \begin{split}
            C_{plan}(p=\{(\pi_1, \theta_1), \ldots, &(\pi_N, \theta_N)\}) = \\ &\sum_{s\in S} Z_{N-1}(s; p)\mathcal{A}_{\pi_N, \theta_N}(s)
    \end{split}
\end{align}
Next we show how to construct plans towards a goal $g \in G$. The key idea is to reinterpret $g$ using affordance. Recall that $g$ is a binary function on whether a state belongs to its goal state set $S_g$. We say that a plan $p=\{(\pi_i, \theta_i)\}_{i=1}^N$ is directed towards goal $g$ if the last skill in the plan can be executed in a goal state, i.e., $A_N \subseteq S_g$. 

\textbf{Definition 2} (Goal-directed plans $\mathcal{P}_g$): \textit{Given a goal $g \in G$ and its goal state set $S_g$, we define the goal-directed plans of $g$ as $\mathcal{P}_g \subseteq \mathcal{P}$ such that $\forall \{(\pi_i, \theta_i)\}_{i=1}^N \in \mathcal{P}_g, \mathcal{A}_{\pi_N, \theta_N}\subseteq S_g$}.

Finally, the problem of searching for a best skill plan towards goal $g \in G$ is:
\begin{align}
\label{eq:optimal-plan}
{\arg\max}_{p\in \mathcal{P}_g} C_{plan}(p)
\end{align}
While it is possible to find exact optimal solutions by computing Eq.~\eqref{eq:optimal-plan} from state space $S$ and transition function $T$, we aim at realistic domains in which we have access to neither. In the following, we present a method that learns to plan in an unknown environment by modeling affordances.

\subsection{\algoNameFull (\algoName)}

We base our learning-to-plan method on a model-based reinforcement learning (MBRL) formulation. To behave in an environment with unknown dynamics, an MBRL agent needs to learn both a dynamics model and a cost function to predict plan-induced future states and evaluate plan costs. Learning the complete model of a large environment in observation space (e.g., predicting future images) is still an open challenge~\cite{oh2015action,agrawal2016learning,finn2017deep,ebert2018visual}. A more effective approach is to build \emph{partial models}~\cite{talvitie2009simple} based on rewards or some task-relevant quantities~\cite{dosovitskiy2016learning,oh2017value,amos2018learning,hafner2019learning}. For example, PlaNet~\cite{hafner2019learning} combines the dynamics and cost modeling by predicting multi-step future rewards through a learned dynamics model. 

Our method can be viewed as building a partial model of the environment based on affordances. Comparing to prior works that rely on task-specific quantities such as rewards, our affordance representation is \emph{task-agnostic}: a grasping skill is afforded regardless of the task goal. This allows our method to share learned affordance models among plans with different goals, comparing favorably to standard MBRL methods that must learn different cost estimates per goal. Affordance models also offer better estimates of plan costs for model-predictive control (described next) compared to methods that rely on estimating the final reward for planning.

Concretely, we jointly train a latent dynamics model and an affordance prediction model. The latent dynamics model predicts future latent states conditioning on sampled skill plans. The affordance model evaluates skill affordances both at the current and future predicted latent states. We use model-predictive control (MPC) to plan in the learned latent space and evaluate the proposed plans by estimating plan completion probabilities (Eq.~\eqref{eq:optimal-plan}) from the predicted affordances. Given a set of parameterized motor skills and their affordance functions, our method iteratively collects data from environment using planning and trains the two models on the gathered data. Here we describe the components. The full iterative learning algorithm is Algorithm~\ref{alg:daf}.

\textbf{Latent dynamics model.} We consider experience sequences $\{(o_t, \pi_t, \theta_t, a_t)\}_{t=1}^{T}$, environment observation $o_t$, a skill $(\pi_i, \theta_i)$ that the robot attempted to execute at time $t$, and the resulting binary affordance value $a_t$. Following PlaNet~\cite{hafner2019learning}, we project observation $o_t$ to a latent encoding $z_t$ using an observation encoder $z_t=f_{enc}(o_t)$. The encoder can be a multi-layer perception for low-dimensional observations and deep CNN for image observations. We make a simplified assumption that the latent dynamics is deterministic~\cite{buesing2018learning} and construct a deterministic transition model $\hat{z}_{t+1}=f_{trans}(z_t, \pi_t, \theta_t)$. We also explore a recurrent transition model $h_{t+1}=f_{trans}(h_t, z_t, \pi_t, \theta_t)$ with decoder $\hat{z}_{t+1}=f_{dec}(h_{t+1})$ that shows better empirical performance on long-horizon tasks.
\input{algo2}

\textbf{Learning dynamics by predicting affordances.} Given latent experiences $\{(z_t, \pi_t, \theta_t, a_t)\}_{t=1}^T$, we train a binary classifier $\hat{a}_t=f_{\mathcal{A}}(z_t,  \pi_t, \theta_t)$ to predict whether a latent state $z_t$ affords the skill $(\pi_t, \theta_t)$. We train the affordance model jointly with the latent dynamics model. The simplest way is to learn from one-step transitions: predicting $\hat{a}_t$ and $\hat{a}_{t+1}$ from $(z_t, \pi_t, \theta_t)$ and $(f_{trans}(z_t, \pi_t, \theta_t), \pi_{t+1}, \theta_{t+1})$, respectively. However, as shown in~\cite{amos2018learning,hafner2019learning}, the latent dynamics model learned from one-step transitions is often not accurate enough for long-horizon planning. Hence we adopt the \emph{overshooting}~\cite{amos2018learning} technique and optimize $f_{trans}$ and $f_{\mathcal{A}}$ over multi-step affordance predictions. 

\textbf{Evaluating plans with approximate models.} Given a pair of learned deterministic dynamics model $f_{trans}$ and affordance model $f_{\mathcal{A}}$, the approximate form of Eq.~\eqref{eq:cost} with respect to an initial latent state $z_1$ is:
\begin{align}
    \label{eq:approx-cost}
    \begin{split}
            \hat{C}_{plan}(p=\{(\pi_i, \theta_i)\}_{i=1}^N) = \prod_{i=1}^N f_{\mathcal{A}}(z_i, \pi_i, \theta_i)  
    \end{split}
\end{align}
where $z_{i} = f_{trans}(z_{i-1}, \theta_{i-1}, \pi_{i-1})$ for $i > 1$.

\textbf{Planning and execution with MPC. } We use a standard model-predictive control (MPC) strategy to plan and execute skills with the learned latent dynamics and affordance models. Given a goal $g$, the MPC planner optimizes:
\begin{align}
    \label{eq:mpc}
    \arg\min_{\pi, \theta} C(\{(z_i, \pi_i, \theta_i)\}_{i=1}^H),
\end{align}
 with plan cost function $C$, goal-directed plans $(\pi_{1:H}, \theta_{1:H}) \in \mathcal{P}_g$, and the latent sequences $z_{1:H}$ generated by $f_{trans}$ over a planning horizon $H$. We use the negative of the approximate plan completion probability in Eq.~\eqref{eq:approx-cost} as the plan cost.

The hybrid discrete-continuous skill plans present a large search space. To accelerate planning, we additionally train a skill skeleton proposal model $\pi_t\sim f_{\pi}(z_t)$ that captures the distribution of feasible discrete skill choices given a latent state $z_t$. Given the sampled discrete skills, we use a na\"ive sampling-based strategy to optimize Eq.~\eqref{eq:mpc} over the continuous skill parameter spaces. We defer more sophisticated strategies such as CEM~\cite{rubinstein1997optimization,hafner2019learning} and gradient-based optimizations~\cite{amos2018learning} to future work. Our agent executes plans using receding-horizon control: execute the first skill in a chosen plan and then replan from the new observation. We present the full planning algorithm in Algorithm~\ref{alg:plan}.

%% file: algo1.tex
\begin{algorithm}
\caption{\textsc{Deep Affordance Foresight}}
\label{alg:daf}
\begin{algorithmic}
\State \textbf{hyperparameters:} \\
Batch size $B$ \\
Number of training iterations $K$ \\
Number of rollouts $R$ \\
Task horizon $T$ \\
Overshooting length $H$
\State \textbf{inputs:} \\
$z=f_{enc}(o)$ \Comment{observation encoder}\\
$\hat{z}_{t+1}=f_{trans}(z_t, \pi_t, \theta_t)$ \Comment{latent transition model}\\
$\hat{a}_t=f_{\mathcal{A}}(z_t,  \pi_t, \theta_t)$ \Comment{affordance model} \\
$\pi_t\sim f_{\pi}(z_t)$ \Comment{skill skeleton proposal model} \\
$\mathcal{D}$ \Comment{replay buffer with seeding data} \\
\texttt{env} \Comment{environment}
\While{\emph{Not Converged}}
\State \verb!// Model fitting!
\For{update steps $k\leftarrow[1...K]$}
    \State Sample experience $\{o^i, \pi^i_{1:H}, \theta^i_{1:H}, a^i_{1:H}\}_{i=1}^B\sim \mathcal{D}$
    \State \verb!// Omitting batch index for clarity!
    \State $z_1\leftarrow f_{enc}(o)$ 
    \For{$h\leftarrow[1...(H - 1)]$}
        \State $z_{h + 1} \leftarrow f_{trans}(z_h, \pi_h, \theta_h)$
    \EndFor
    \State Train $f_{\mathcal{A}}(z_{1:H}, \pi_{1:H}, \theta_{1:H})$ against affordance labels $a_{1:H}$ with BCE loss.
    \State Train $f_\pi(z_{1:H})$ against skills that were executed with discrete-distribution MLE.
\EndFor

\State \verb!// Experience collection!
\For{rollout iteration $r\leftarrow[1...R]$}
    \State $o, g \leftarrow \texttt{env.reset()}$ \Comment{get obs $o$ and goal $g$}
    \For{rollout step $t\leftarrow[1...T]$}
        \State $\pi, \theta \leftarrow$\textsc{PlanWithAffordance} \Comment{plan for $g$}
        \State $a \leftarrow\texttt{env.skill\_is\_executable}(\pi, \theta)$
        \If{$a = 1$}
            \State $o\leftarrow \texttt{env.step}(\pi, \theta)$
        \EndIf
        \State Append experience $(o, \pi, \theta, a)$ to $\mathcal{D}$
    \EndFor
\EndFor
\EndWhile

\end{algorithmic}
\end{algorithm}

%% file: algo2.tex

\begin{algorithm}
\caption{\textsc{PlanWithAffordance}}
\label{alg:plan}
\begin{algorithmic}
\State \textbf{hyperparameters:} planning horizon $H$, num. samples $N$
\State \textbf{inputs:} \\
$z=f_{enc}(o)$ \Comment{observation encoder}\\
$\hat{z}_{t+1}=f_{trans}(z_t, \pi_t, \theta_t)$ \Comment{latent transition model}\\
$\hat{a}_t=f_{\mathcal{A}}(z_t,  \pi_t, \theta_t)$ \Comment{affordance model} \\
$\pi_t\sim f_{\pi}(z_t)$ \Comment{skill skeleton proposal model} \\
$o$ \Comment{current environment observation} \\
$\theta \sim param(\pi)$ \Comment{random skill parameter sampler} \\
$P_g$ \Comment{set of goal-directed plans for goal $g$}

\State \textbf{start}
\State \texttt{plans} $\leftarrow []$ \Comment{sampled plans}
\State \texttt{affs} $\leftarrow []$ \Comment{step-wise affordances}
\State $z_1\leftarrow f_{enc}(o)$ \Comment{encode observation to latent}
\State $z^{1:N}_1 \leftarrow repeat(z_1, N)$ \Comment{repeat latent $N$ times}

\State \verb!// Shooting method with sampled skills!
\For{$i \leftarrow [1, ..., H]$}
\State  $\pi^{1:N}_i \sim f_{\pi}(z^{1:N}_i)$ \Comment{sample skill skeletons}
\State $\theta^{1:N}_i \sim param(\pi^{1:N}_i)$ \Comment{sample parameters}
\State $a^{1:N}_i = f_{\mathcal{A}}(z_i, \pi^{1:N}_i, \theta^{1:N}_i)$ \Comment{compute affordance}
\State $\texttt{plans} \leftarrow \texttt{plans} \cup (\pi^{1:N}_i, \theta^{1:N}_i)$ 
\State $\texttt{affs} \leftarrow \texttt{affs} \cup a^{1:N}_i$ 
\State $z^{1:N}_{i + 1} \leftarrow f_{trans}(z^{1:N}_i, \pi^{1:N}_i, \theta^{1:N}_i)$ \Comment{forward dynamics}
\EndFor
\State \verb!// Evaluate plan cost using affordances!
\For{$k \leftarrow [1, ..., N]$}
\State $(\pi_{1:H}, \theta_{1:H}) \leftarrow \texttt{plans}[k]$ \Comment{$k$-th plan in \texttt{plans}}
\State $a_{1:H} \leftarrow \texttt{affs}[k]$  \Comment{step-wise affordances}
\State$c_k \leftarrow \begin{cases}
-\prod_{t=1}^H a_t \quad &\text{if }(\pi_{1:H}, \theta_{1:H}) \in \mathcal{P}_g\text{ (Eq.~\eqref{eq:approx-cost})}  \\[5pt]
\infty\quad &\text{otherwise}
\end{cases}$
\EndFor

\State $k \leftarrow \arg\min_{k=\{1...N\}} (c_k)$ \Comment{get the lowest-cost plan}
\State $\pi^{*}, \theta^{*} \leftarrow \texttt{plans}[k][0]$ \Comment{first skill of the chosen plan}
\State \Return $\pi^{*}, \theta^{*}$ 
\end{algorithmic}
\end{algorithm}

%% file: 4_exp.tex
\section{Experiments}



\begin{figure*}[t]
  \centering
  \includegraphics[width=0.9\linewidth]{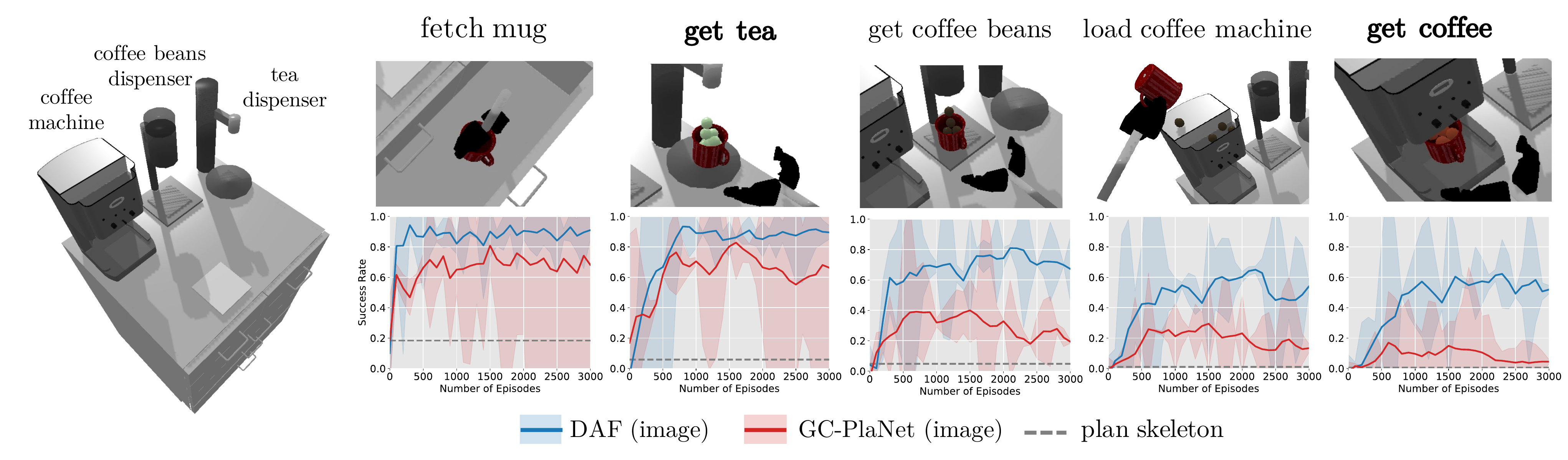}
  \caption{Setup (left) and results on key stages of the \emph{Kitchen} task domain, which features challenging dynamics such as liquid-like objects (load coffee machine). We compare our method \algoName and GC-PlaNet on learning to achieve the two final goals, \emph{get coffee} and \emph{get tea}, with raw image input. 
  }
  \vspace{-10pt}
  \label{fig:kitchen}
\end{figure*}

Our experiments seek to validate the primary claims that (1) our affordance representation supports long-horizon planning of challenging manipulation tasks, (2) our method can outperform reward-based latent planning methods~\cite{hafner2019learning}, (3) the task-agnostic latent transition and affordance models facilitate knowledge sharing among multiple tasks, and (4) our method can plan directly with raw image input. We conduct evaluations in two simulated environments: a \emph{Tool-Use} environment (Fig.~\ref{fig:pull}) for analyzing the key traits of our method and a \emph{Kitchen} (Fig.~\ref{fig:kitchen}) environment that features complex visual scenes and tasks that require handling non-rigid dynamics of liquid-like objects.

\subsection{Tool-Use Domain}
\textbf{Task setup } The task design is primarily inspired by \cite{toussaint2018differentiable,loula2020development}. The environments as shown in Fig.~\ref{fig:pull} are simulated using PyBullet~\cite{BulletPhysics}. We evaluate on two sets of tasks in this domain. The first is \textbf{tool-use} with two task goals: use the tool to fetch and place either the blue or the red cube on the green target respectively. A virtual wall prevents the gripper from directly grasping the red cube - the robot must use the tool to pull it across the wall. The blue cube is in a pipe - the robot needs to use the tool to push it out of the pipe before being able to grasp it. The two task goals are sampled randomly each episode. \textbf{tool-use + stack} is a longer task of stacking the two cubes on top of the target, which requires the robot to use the tool differently to fetch both cubes.

\textbf{Parameterized skills } The agent is provided with four parameterized motor skills: \texttt{grasp}, \texttt{place}, \texttt{hook}, and \texttt{poke}. \texttt{grasp} executes top-down grasps parameterized by 3D grasping locations relative to the target object. \texttt{place} sets a grasped object onto a surface parameterized by the relative 2D location between the object and the surface. Both \texttt{hook} and \texttt{poke} moves the object-in-hand along a trajectory with parameterized start and end positions. The motion trajectories are generated using RRT-based~\cite{kuffner2000rrt} motion planners. We use additional ``no-op'' skills to specify goals. no-op skills are skills that have affordance sets but do not incur changes to the environment if executed. Each goal (\emph{red-on-target, blue-on-target, stack-red-blue-on-target}) is associated with a no-op \emph{goal skill} for which the affordance set is equal to the goal state set. In other words, the goal skills are only afforded at their corresponding goal states. 

\textbf{Skill feasibilities } Skill feasibilities are determined by pre-defined workspace constraints (e.g., the gripper cannot go beyond the virtual wall) and PyBullet's built-in collision detector for checking if a motion plan would cause unintended collisions between the robot and the environment. For example, we consider a grasping skill that would result in the gripper colliding with the table as infeasible. Conversely, grasping skills that do not touch any object at all are considered to be feasible. In the real world, collision detection can be implemented through a depth-based octomap~\cite{chitta2012moveit}.

\textbf{Architectures and baselines } To isolate the effects of our method and design choices, we focus on object pose input space in this domain. We evaluate our method with recurrent dynamics (\textbf{\algoName}) and MLP-based dynamics (\textbf{\algoName (no RNN)}). All other components, $f_{enc}$, $f_{\mathcal{A}}$, and $f_{trans}$, are MLPs. We compare with a goal-conditional variant of PlaNet~\cite{hafner2019learning} (\textbf{GC-PlaNet}) by conditioning the learned reward model on a task ID. To facilitate fair comparisons, we remove the auxiliary observation model and the stochastic component in the recurrent dynamics of PlaNet, and match all other architecture choices to \algoName. These components are orthogonal to the comparison and adding them to our framework will be explored in future work. We also include a hard-coded baseline (\textbf{plan skeleton}) that executes ground truth plan skeletons (discrete skills) with random skill parameters in open-loop to highlight that the tasks we consider require intelligent skill parameter selection in conjunction with the correct skill sequence to solve consistently. 

\begin{figure}
  \centering
  \includegraphics[width=1.0\linewidth]{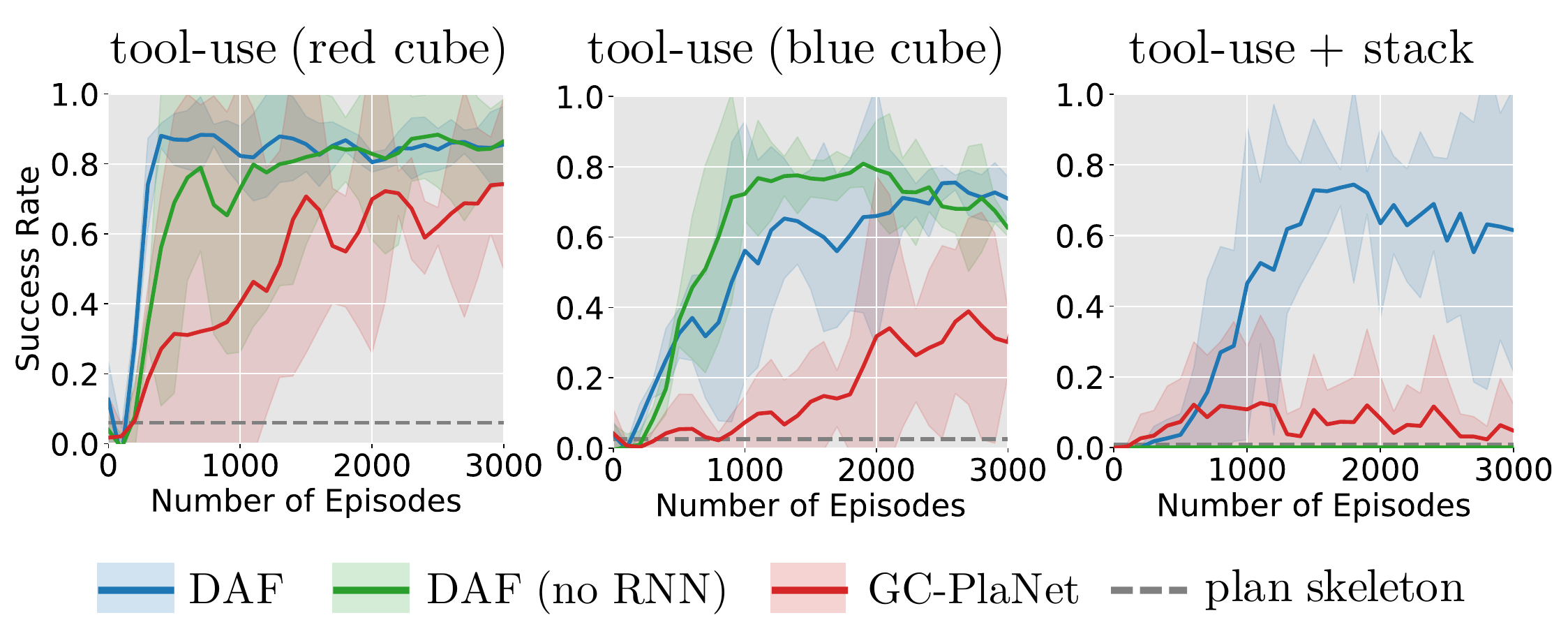}
     \vspace{-15pt}
  \caption{Results on the jointly learning the two \emph{tool-use} tasks shown in Fig.~\ref{fig:pull} and a combined \emph{tool-use + stack} task, where the robot has to use the tool to get both cubes and stack them on the green target. We compare our method (\algoName) and a goal-conditional variant of PlaNet~\cite{hafner2019learning} (GC-PlaNet). 
  }
  \label{fig:tool-base}
    \vspace{-10pt}
\end{figure}

\begin{figure}
  \centering
  \includegraphics[width=1.0\linewidth]{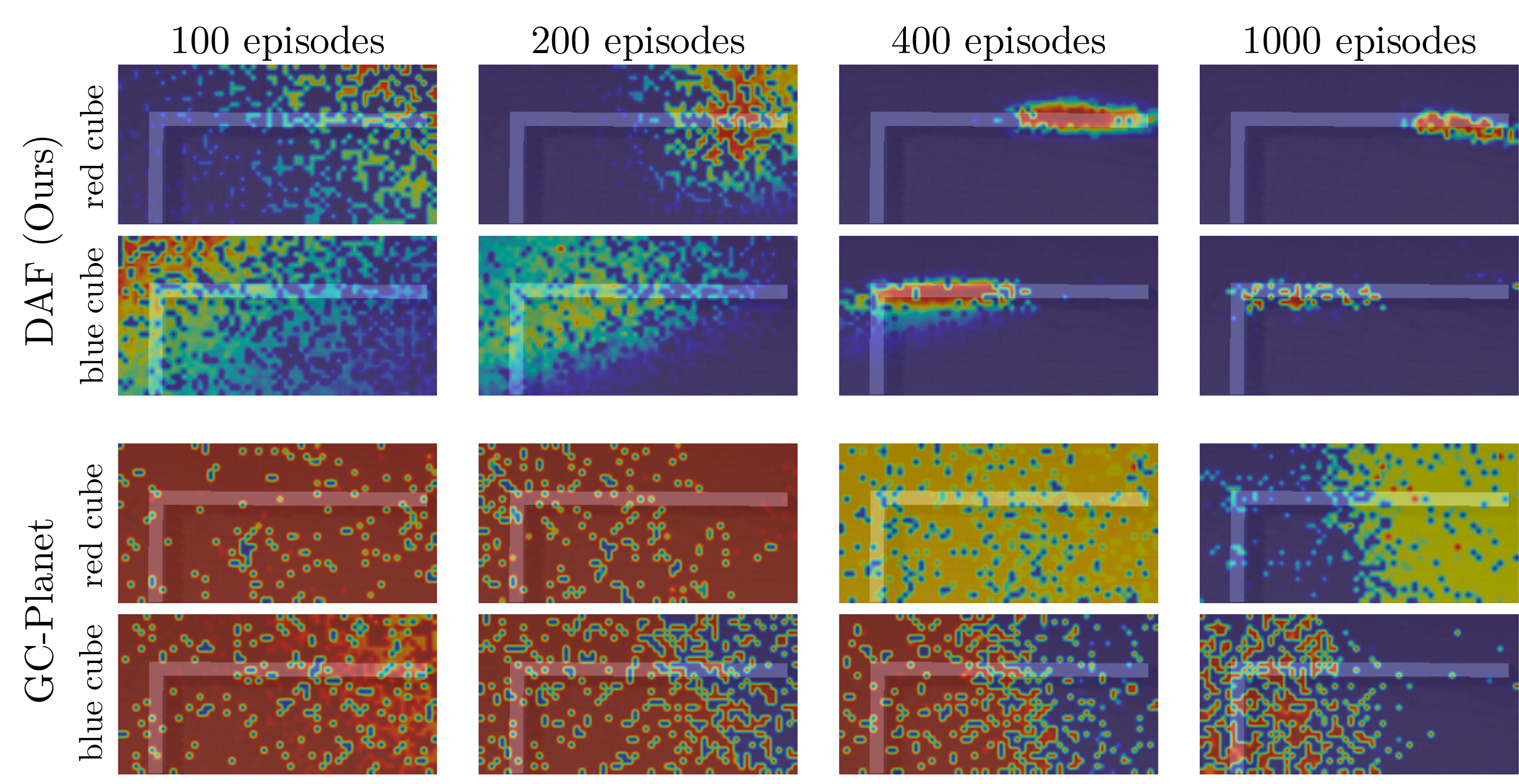}
  \caption{Visualizing the plan score predictions over the course of learning the two \emph{tool-use} tasks. Each column shows the prediction made by models trained with $N$ number of actively collected episodes. Each pixel of the heat map shows the predicted score (Eq.\eqref{eq:approx-cost}) of a plan that starts with the \texttt{grasp} skill parameterized by the corresponding $x, y$ location.
  }
  \label{fig:tool-qual}
  \vspace{-10pt}
\end{figure}

\textbf{Results } Left two plots in Fig.~\ref{fig:tool-base} show the results of jointly learning the two \emph{tool-use} tasks.  We see that \algoName converges to high success rate within 1000 episodes for both tasks. GC-PlaNet performs competitively on the red-cube task but peaks at 0.4 success rate for the blue cube task, which requires more careful grasping pose choice for poking the blue cube out of the pipe. 
The rightmost plot in Fig.~\ref{fig:tool-base} shows the results on a longer task that requires the robot to fetch the two cubes and stack them on the green target.  On the left figure, we observe that \algoName reaches peak performance of 0.7 success rate at episode 1500, whereas GC-PlaNet converges at 0.1 success rate. Notably, \algoName without RNN dynamics falls flat on this task, echoing the findings in \cite{hafner2019learning,amos2018learning} that recurrent dynamics is crucial for modeling long tasks.


\textbf{Analyzing learned affordances } To get a better idea of how the affordance representations learned by \algoName develop over the training process, we visualize the plan scores computed from the learned affordance values using Eq.\eqref{eq:approx-cost}. Specifically, we visualize plan scores at the beginning of both \emph{tool-use} tasks, since both task goals require the robot to grasp the tool. As shown in Fig.~\ref{fig:tool-qual}, each pixel in the overlaid heatmap (red-high, blue-low) indicates the normalized score of a plan that starts with a \texttt{grasp} skill parameterized by the corresponding $x, y$ location with a constant $z$-height. Each column shows the visualization produced by models trained with certain number of actively collected episodes. 

We see that \algoName is able to rapidly learn meaningful affordance representations with respect to each task goal with as little as 400 actively collected episodes. In contrast, GC-PlaNet's plan score prediction remains noisy even at 1000 episodes. One may also notice that \algoName continues to shrink its ``good grasp'' predictions at episodes 1000. This is because while the plan scores are only used to decide the next skill to execute, they are computed from the \emph{future skills affordances} over the rest of a plan. As the training progresses, the agent starts to reach later stages of the task and get better estimates of the skill affordances later in the plan, which will in turn influence the plan scores even at the beginning of an episode. This behavior highlights the key difference between our future-aware affordance representation and traditional affordances that only model myopic effects of actions.

\subsection{Kitchen Domain}
Compared to TAMP-like methods that require a hand-defined planning space (e.g. object poses), our method can learn end-to-end with raw image inputs. This allows our method to learn to plan through complex non-rigid dynamics such as pouring liquid. To test this capability, we task the robot to serve tea and coffee in the \emph{Kitchen} domain as shown on the left side of Fig.\ref{fig:kitchen} with only visual inputs.

\textbf{Setup } The domain has two tasks of varying difficulties: in the simpler \textbf{get tea} task, the robot needs to open the drawer, fetch the mug, use the platform to reorient the grasp and set the mug at the correct location beneath the tea dispenser tap to get tea. In a more challenging \textbf{get coffee} task, the robot needs to fetch the mug from the drawer, use the mug to get the coffee beans from the dispenser, then pour the coffee beans into the coffee machine, and finally set the mug beneath the coffee machine dispenser to get coffee. The two goals are sampled randomly each episode. We use small spherical beads to approximate liquid dynamics.

The robot is equipped with the following parameterized skills: \texttt{grasp} is parameterized with gripper-object distance and two discrete grasping orientations: side and top; \texttt{place} with the relative location between the object to be place and a surface object; \texttt{pour} is parameterized by the relative position between the object-in-hand and the target container and a pouring angle; \texttt{open} is parameterized by a grasp location and a distance to pull along a given direction. We use no-op to skills to represent the goals.


The environment observations are RGB images rendered at $128\times 128$ resolution from the perspective shown in Fig.~\ref{fig:kitchen}. Accordingly, we change $f_{enc}$ to a ResNet architecture~\cite{he2015deep} followed by a a Spatial-Softmax layer~\cite{finn2016deep} and an MLP. The remaining components are the same as in \emph{Tool-Use}.

\textbf{Results } As shown in Fig.~\ref{fig:kitchen}, \algoName is able to jointly learn both tasks, get tea and get coffee, with high success rate from only raw image inputs. Moreover, we observe that both \algoName and GC-PlaNet can solve the simpler \emph{get tea} tasks, with \algoName having significantly lower performance variance. For the more challenging \emph{get coffee} task, \algoName learns to fill the mug with coffee beans within 500 episodes and learns to get coffee from the coffee machine at 0.6 success rate in 1500 episodes, whereas GC-PlaNet plateaus at $<$0.2 success rate. 

\begin{SCfigure}
  \centering
  \includegraphics[height=4cm]{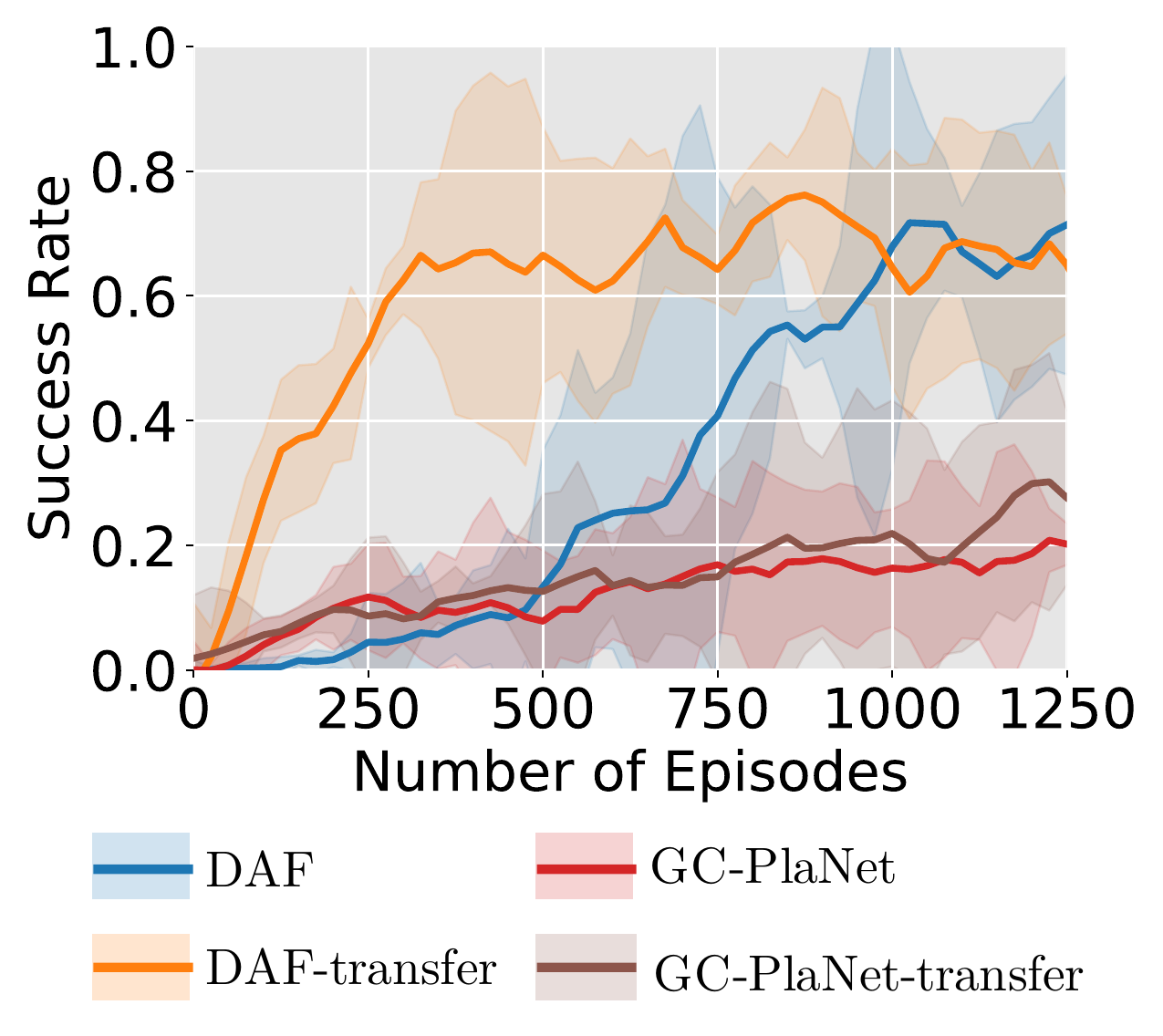}  
  \caption{We compare learning the standalone \emph{get coffee} task from scratch vs. finetuning from models pre-trained on the \emph{get tea} task. Our task-agnostic affordance representation allows \algoName to transfer knowledge about how to fetch the mug from the short get-tea task to the longer get-coffee, enabling \algoName to learn a performant policy within 300 episodes.}
  \label{fig:finetune}
  \vspace{-5pt}
\end{SCfigure}

We in addition highlight that the task-agnostic affordance representation allows \algoName to share the learned affordance and latent dynamics models across tasks. To verify this claim, we compare learning the standalone \emph{get coffee} task a) from scratch and b) finetuning from models \emph{pretrained} on the \emph{get tea} task. \algoName should be able to transfer the affordances for opening the drawer and fetching the mug from the short \emph{get tea} task to the longer \emph{get coffee} task. To remove conflating factors such as image encoders, we evaluate the models on a hand-defined feature space of object poses and the number of \{coffee, coffee bean, tea\} beads contained in each object. As Fig.~\ref{fig:finetune} shows, \algoName pretrained with the \emph{get tea} task is able to learn the \emph{get coffee} task with only 300 actively collected episodes. In contrast, the pretrained GC-PlaNet shows no significant improvement compared to learning from scratch.

%% file: 5_conclusion.tex
\section{Conclusions}
In this paper, we (1) extended the classical concept of affordance to multi-step robot planning and (2) presented a learning-to-plan framework to interactively build environment models around skill affordances and to solve multi-step tasks via trial and error. In two challenging manipulation domains, we show that our method outperforms existing latent-space planning method, is able to reuse the task-agnostic affordance representation to adapt to a new longer task, and learn to plan with high-dimensional image input.

%% file: refs.bib
@inproceedings{oh2017value,
  title={Value prediction network},
  author={Oh, Junhyuk and Singh, Satinder and Lee, Honglak},
  booktitle={Advances in Neural Information Processing Systems},
  pages={6118--6128},
  year={2017}
}

@inproceedings{agrawal2016learning,
  title={Learning to poke by poking: Experiential learning of intuitive physics},
  author={Agrawal, Pulkit and Nair, Ashvin V and Abbeel, Pieter and Malik, Jitendra and Levine, Sergey},
  booktitle={Advances in Neural Information Processing Systems},
  pages={5074--5082},
  year={2016}
}

@inproceedings{konidaris2009skill,
  title={Skill discovery in continuous reinforcement learning domains using skill chaining},
  author={Konidaris, George and Barto, Andrew G},
  booktitle={Advances in neural information processing systems},
  pages={1015--1023},
  year={2009}
}

@article{james2019learning,
  title={Learning Portable Representations for High-Level Planning},
  author={James, Steven and Rosman, Benjamin and Konidaris, George},
  journal={ICML},
  year={2020}
}

@article{konidaris2012robot,
  title={Robot learning from demonstration by constructing skill trees},
  author={Konidaris, George and Kuindersma, Scott and Grupen, Roderic and Barto, Andrew},
  journal={The International Journal of Robotics Research},
  volume={31},
  number={3},
  pages={360--375},
  year={2012},
  publisher={SAGE Publications Sage UK: London, England}
}

@article{xia2018learning,
  title={Learning sparse relational transition models},
  author={Xia, Victoria and Wang, Zi and Kaelbling, Leslie Pack},
  journal={International Conference on Learning Representations},
  year={2018}
}

@article{he2015deep,
  title={Deep residual learning for image recognition},
  author={He, Kaiming and Zhang, Xiangyu and Ren, Shaoqing and Sun, Jian},
  journal={CVPR},
  year={2016}
}

@article{mahler2017dexnet,
  title={Dex-Net 2.0: Deep Learning to Plan Robust Grasps with Synthetic Point Clouds and Analytic Grasp Metrics},
  author={Jeffrey Mahler and Jacky Liang and Sherdil Niyaz and Michael Laskey and Richard Doan and Xinyu Liu and Juan Aparicio Ojea and Ken Goldberg},
  journal={arXiv preprint arXiv:1703.09312},
  year={2017}
}

@article{dosovitskiy2016learning,
  title={Learning to act by predicting the future},
  author={Dosovitskiy, Alexey and Koltun, Vladlen},
  journal={ICLR},
  year={2017}
}

@article{fang2020learning,
  title={Learning task-oriented grasping for tool manipulation from simulated self-supervision},
  author={Fang, Kuan and Zhu, Yuke and Garg, Animesh and Kurenkov, Andrey and Mehta, Viraj and Fei-Fei, Li and Savarese, Silvio},
  journal={The International Journal of Robotics Research},
  volume={39},
  number={2-3},
  pages={202--216},
  year={2020},
  publisher={SAGE Publications Sage UK: London, England}
}

@inproceedings{song2010learning,
  title={Learning task constraints for robot grasping using graphical models},
  author={Song, Dan and Huebner, Kai and Kyrki, Ville and Kragic, Danica},
  booktitle={2010 IEEE/RSJ International Conference on Intelligent Robots and Systems},
  pages={1579--1585},
  year={2010},
  organization={IEEE}
}

@inproceedings{zeng2018learning,
  title={Learning synergies between pushing and grasping with self-supervised deep reinforcement learning},
  author={Zeng, Andy and Song, Shuran and Welker, Stefan and Lee, Johnny and Rodriguez, Alberto and Funkhouser, Thomas},
  booktitle={2018 IEEE/RSJ International Conference on Intelligent Robots and Systems (IROS)},
  pages={4238--4245},
  year={2018},
  organization={IEEE}
}

@inproceedings{dang2012semantic,
  title={Semantic grasping: Planning robotic grasps functionally suitable for an object manipulation task},
  author={Dang, Hao and Allen, Peter K},
  booktitle={2012 IEEE/RSJ International Conference on Intelligent Robots and Systems},
  pages={1311--1317},
  year={2012},
  organization={IEEE}
}

@inproceedings{abel2014toward,
  title={Toward affordance-aware planning},
  author={Abel, David and Barth-Maron, Gabriel and MacGlashan, James and Tellex, Stefanie},
  booktitle={First Workshop on Affordances: Affordances in Vision for Cognitive Robotics},
  year={2014}
}

@article{cruz2016training,
  title={Training agents with interactive reinforcement learning and contextual affordances},
  author={Cruz, Francisco and Magg, Sven and Weber, Cornelius and Wermter, Stefan},
  journal={IEEE Transactions on Cognitive and Developmental Systems},
  volume={8},
  number={4},
  pages={271--284},
  year={2016},
  publisher={IEEE}
}

@inproceedings{finn2016deep,
  title={Deep spatial autoencoders for visuomotor learning},
  author={Finn, Chelsea and Tan, Xin Yu and Duan, Yan and Darrell, Trevor and Levine, Sergey and Abbeel, Pieter},
  booktitle={2016 IEEE International Conference on Robotics and Automation (ICRA)},
  pages={512--519},
  year={2016},
  organization={IEEE}
}

@inproceedings{kuffner2000rrt,
  title={RRT-connect: An efficient approach to single-query path planning},
  author={Kuffner, James J and LaValle, Steven M},
  booktitle={Proceedings 2000 ICRA. Millennium Conference. IEEE International Conference on Robotics and Automation. Symposia Proceedings (Cat. No. 00CH37065)},
  volume={2},
  pages={995--1001},
  year={2000},
  organization={IEEE}
}

@article{konidaris2014constructing,
  title={Constructing symbolic representations for high-level planning},
  author={Konidaris, George D and Kaelbling, Leslie P and Lozano-Perez, Tomas},
  year={2014},
  publisher={Association for the Advancement of Artificial Intelligence (AAAI)}
}

@inproceedings{konidaris2015symbol,
  title={Symbol acquisition for probabilistic high-level planning},
  author={Konidaris, George and Kaelbling, Leslie P and Lozano-Perez, Tomas},
  year={2015},
  organization={AAAI Press/International Joint Conferences on Artificial Intelligence}
}

@article{konidaris2018skills,
  title={From skills to symbols: Learning symbolic representations for abstract high-level planning},
  author={Konidaris, George and Kaelbling, Leslie Pack and Lozano-Perez, Tomas},
  journal={Journal of Artificial Intelligence Research},
  volume={61},
  pages={215--289},
  year={2018}
}

@article{chitta2012moveit,
  title={Moveit![ros topics]},
  author={Chitta, Sachin and Sucan, Ioan and Cousins, Steve},
  journal={IEEE Robotics \& Automation Magazine},
  volume={19},
  number={1},
  pages={18--19},
  year={2012},
  publisher={IEEE}
}

@inproceedings{abel2015goal,
  title={Goal-based action priors},
  author={Abel, David and Hershkowitz, David Ellis and Barth-Maron, Gabriel and Brawner, Stephen and O'Farrell, Kevin and MacGlashan, James and Tellex, Stefanie},
  booktitle={Twenty-Fifth International Conference on Automated Planning and Scheduling},
  year={2015}
}

@article{detry2011learning,
  title={Learning grasp affordance densities},
  author={Detry, Renaud and Kraft, Dirk and Kroemer, Oliver and Bodenhagen, Leon and Peters, Jan and Kr{\"u}ger, Norbert and Piater, Justus},
  journal={Paladyn, Journal of Behavioral Robotics},
  volume={2},
  number={1},
  pages={1--17},
  year={2011},
  publisher={De Gruyter}
}

@inproceedings{ugur2007learning,
  title={The learning and use of traversability affordance using range images on a mobile robot},
  author={Ugur, Emre and Dogar, Mehmet R and Cakmak, Maya and Sahin, Erol},
  booktitle={Proceedings 2007 IEEE International Conference on Robotics and Automation},
  pages={1721--1726},
  year={2007},
  organization={IEEE}
}

@article{nagarajan2020learning,
  title={Learning Affordance Landscapes forInteraction Exploration in 3D Environments},
  author={Nagarajan, Tushar and Grauman, Kristen},
  journal={arXiv preprint arXiv:2008.09241},
  year={2020}
}

@article{mandikal2020dexterous,
  title={Dexterous Robotic Grasping with Object-Centric Visual Affordances},
  author={Mandikal, Priyanka and Grauman, Kristen},
  journal={arXiv preprint arXiv:2009.01439},
  year={2020}
}

@inproceedings{bousmalis2018using,
  title={Using simulation and domain adaptation to improve efficiency of deep robotic grasping},
  author={Bousmalis, Konstantinos and Irpan, Alex and Wohlhart, Paul and Bai, Yunfei and Kelcey, Matthew and Kalakrishnan, Mrinal and Downs, Laura and Ibarz, Julian and Pastor, Peter and Konolige, Kurt and others},
  booktitle={2018 IEEE international conference on robotics and automation (ICRA)},
  pages={4243--4250},
  year={2018},
  organization={IEEE}
}

@article{buesing2018learning,
  title={Learning and querying fast generative models for reinforcement learning},
  author={Buesing, Lars and Weber, Theophane and Racaniere, S{\'e}bastien and Eslami, SM and Rezende, Danilo and Reichert, David P and Viola, Fabio and Besse, Frederic and Gregor, Karol and Hassabis, Demis and others},
  journal={arXiv preprint arXiv:1802.03006},
  year={2018}
}

@article{amos2018learning,
  title={Learning awareness models},
  author={Amos, Brandon and Dinh, Laurent and Cabi, Serkan and Roth{\"o}rl, Thomas and Colmenarejo, Sergio G{\'o}mez and Muldal, Alistair and Erez, Tom and Tassa, Yuval and de Freitas, Nando and Denil, Misha},
  journal={International Conference on Learning Representations},
  year={2018}
}

@article{rubinstein1997optimization,
  title={Optimization of computer simulation models with rare events},
  author={Rubinstein, Reuven Y},
  journal={European Journal of Operational Research},
  volume={99},
  number={1},
  pages={89--112},
  year={1997},
  publisher={Elsevier}
}

@inproceedings{ames2018learning,
  title={Learning symbolic representations for planning with parameterized skills},
  author={Ames, Barrett and Thackston, Allison and Konidaris, George},
  booktitle={2018 IEEE/RSJ International Conference on Intelligent Robots and Systems (IROS)},
  pages={526--533},
  year={2018},
  organization={IEEE}
}

@MISC{BulletPhysics,
author =   {Erwin Coumans and Yunfei Bai},
title =    {pybullet, a Python module for physics simulation, games, robotics and machine learning},
howpublished = {\url{http://pybullet.org/}},
year = {2017}
}

@article{garrett2020integrated,
  title={Integrated Task and Motion Planning},
  author={Garrett, Caelan Reed and Chitnis, Rohan and Holladay, Rachel and Kim, Beomjoon and Silver, Tom and Kaelbling, Leslie Pack and Lozano-P{\'e}rez, Tom{\'a}s},
  journal={arXiv preprint arXiv:2010.01083},
  year={2020}
}

@article{loula2020development,
  title={A Task and Motion Approach to the Development of Planning},
  author={Joao Loula, Kelsey R. Allen, Joshua B. Tenenbaum},
  journal={CogSci},
  year={2020}
}

@inproceedings{kaelbling2011hierarchical,
  title={Hierarchical task and motion planning in the now},
  author={Kaelbling, Leslie Pack and Lozano-P{\'e}rez, Tom{\'a}s},
  booktitle={ICRA},
  year={2011}
}

@incollection{kaelbling2017pre,
  title={Pre-image backchaining in belief space for mobile manipulation},
  author={Kaelbling, Leslie Pack and Lozano-P{\'e}rez, Tom{\'a}s},
  booktitle={Robotics Research},
  pages={383--400},
  year={2017},
  publisher={Springer}
}

@inproceedings{hafner2019learning,
  title={Learning latent dynamics for planning from pixels},
  author={Hafner, Danijar and Lillicrap, Timothy and Fischer, Ian and Villegas, Ruben and Ha, David and Lee, Honglak and Davidson, James},
  booktitle={International Conference on Machine Learning},
  pages={2555--2565},
  year={2019},
  organization={PMLR}
}

@article{kaelbling2013integrated,
  title={Integrated task and motion planning in belief space},
  author={Kaelbling, Leslie Pack and Lozano-P{\'e}rez, Tom{\'a}s},
  journal={The International Journal of Robotics Research},
  volume={32},
  number={9-10},
  pages={1194--1227},
  year={2013},
  publisher={Sage Publications Sage UK: London, England}
}

@inproceedings{kaelbling2017learning,
  title={Learning composable models of parameterized skills},
  author={Kaelbling, Leslie Pack and Lozano-P{\'e}rez, Tom{\'a}s},
  booktitle={2017 IEEE International Conference on Robotics and Automation (ICRA)},
  pages={886--893},
  year={2017},
  organization={IEEE}
}

@article{pasula2007learning,
  title={Learning symbolic models of stochastic domains},
  author={Pasula, Hanna M and Zettlemoyer, Luke S and Kaelbling, Leslie Pack},
  journal={Journal of Artificial Intelligence Research},
  volume={29},
  pages={309--352},
  year={2007}
}

@article{driess2020deepreasoning,
  title={Deep Visual Reasoning: Learning to Predict Action Sequences for Task and Motion Planning from an Initial Scene Image},
  author={Driess, Danny and Ha, Jung-Su and Toussaint, Marc},
  journal={arXiv preprint arXiv:2006.05398},
  year={2020}
}

@inproceedings{driess2020deepheuristics,
  title={Deep visual heuristics: Learning feasibility of mixed-integer programs for manipulation planning},
  author={Driess, Danny and Oguz, Ozgur and Ha, Jung-Su and Toussaint, Marc},
  booktitle={Proc. of the IEEE International Conference on Robotics and Automation (ICRA)},
  year={2020}
}

@article{gibson1977theory,
  title={The theory of affordances},
  author={Gibson, James J},
  journal={Hilldale, USA},
  volume={1},
  number={2},
  year={1977}
}

@article{wells2019learning,
  title={Learning feasibility for task and motion planning in tabletop environments},
  author={Wells, Andrew M and Dantam, Neil T and Shrivastava, Anshumali and Kavraki, Lydia E},
  journal={IEEE robotics and automation letters},
  volume={4},
  number={2},
  pages={1255--1262},
  year={2019},
  publisher={IEEE}
}

@inproceedings{da2014active,
  title={Active learning of parameterized skills},
  author={Da Silva, Bruno and Konidaris, George and Barto, Andrew},
  booktitle={International Conference on Machine Learning},
  pages={1737--1745},
  year={2014}
}

@inproceedings{talvitie2009simple,
  title={Simple local models for complex dynamical systems},
  author={Talvitie, Erik and Singh, Satinder P},
  booktitle={Advances in Neural Information Processing Systems},
  pages={1617--1624},
  year={2009}
}

@article{khetarpalcan,
  title={What can I do here? A Theory of Affordances in Reinforcement Learning},
  author={Khetarpal, Khimya and Ahmed, Zafarali and Comanici, Gheorghe and Abel, David and Precup, Doina},
  booktitle={International Conference on Machine Learning},
  year={2020}
}

@inproceedings{watter2015embed,
  title={Embed to control: A locally linear latent dynamics model for control from raw images},
  author={Watter, Manuel and Springenberg, Jost and Boedecker, Joschka and Riedmiller, Martin},
  booktitle={Advances in neural information processing systems},
  pages={2746--2754},
  year={2015}
}

@inproceedings{finn2017deep,
  title={Deep visual foresight for planning robot motion},
  author={Finn, Chelsea and Levine, Sergey},
  booktitle={ICRA},
  pages={2786--2793},
  year={2017},
  organization={IEEE}
}

@article{ebert2018visual,
  title={Visual foresight: Model-based deep reinforcement learning for vision-based robotic control},
  author={Ebert, Frederik and Finn, Chelsea and Dasari, Sudeep and Xie, Annie and Lee, Alex and Levine, Sergey},
  journal={arXiv preprint arXiv:1812.00568},
  year={2018}
}

@inproceedings{oh2015action,
  title={Action-conditional video prediction using deep networks in {ATARI} games},
  author={Oh, Junhyuk and Guo, Xiaoxiao and Lee, Honglak and Lewis, Richard L and Singh, Satinder},
  booktitle={NIPS},
  pages={2863--2871},
  year={2015}
}

@article{toussaintlogic,
  title={Logic-Geometric Programming: An Optimization-Based Approach to Combined Task and Motion Planning},
  author={Toussaint, Marc}
}

@article{toussaint2018differentiable,
  title={Differentiable physics and stable modes for tool-use and manipulation planning},
  author={Toussaint, Marc A and Allen, Kelsey Rebecca and Smith, Kevin A and Tenenbaum, Joshua B},
  year={2018},
  publisher={Robotics: Science and Systems Foundation}
}

@inproceedings{garrett2020pddlstream,
  title={PDDLStream: Integrating symbolic planners and blackbox samplers via optimistic adaptive planning},
  author={Garrett, Caelan Reed and Lozano-P{\'e}rez, Tom{\'a}s and Kaelbling, Leslie Pack},
  booktitle={Proceedings of the International Conference on Automated Planning and Scheduling},
  volume={30},
  pages={440--448},
  year={2020}
}

@inproceedings{wang2018active,
  title={Active model learning and diverse action sampling for task and motion planning},
  author={Wang, Zi and Garrett, Caelan Reed and Kaelbling, Leslie Pack and Lozano-P{\'e}rez, Tom{\'a}s},
  booktitle={2018 IEEE/RSJ International Conference on Intelligent Robots and Systems (IROS)},
  pages={4107--4114},
  year={2018},
  organization={IEEE}
}
